\documentclass[opre,sglanonrev]{style/informs4_arxiv}
\usepackage{style/eqndefns-left} 
\RequirePackage{tgtermes}
\RequirePackage{newtxtext}
\RequirePackage{newtxmath}
\RequirePackage{bm}
\RequirePackage{endnotes}
\usepackage{amsmath,amsfonts,cancel}
\usepackage{thmtools,enumitem,subfigure}
\usepackage{mathtools}
\usepackage{dsfont}
\usepackage{tcolorbox}
\usepackage{tensor}
\usepackage{array}
\usepackage{caption}
\usepackage{bm}
\usepackage{caption}
\usepackage{hyperref}
\usepackage{placeins}
\usepackage{natbib}
\usepackage{booktabs}
\usepackage{adjustbox}

\mathchardef\mhyphen="2D 

\let\originalleft\left
\let\originalright\right
\renewcommand{\left}{\mathopen{}\mathclose\bgroup\originalleft}
\renewcommand{\right}{\aftergroup\egroup\originalright}

\usepackage{color-edits}
\usepackage{multirow}
\addauthor[Yingxi]{yl}{red}
\addauthor[Anders]{aw}{green}
\addauthor[Ellen]{ev}{magenta}
\OneAndAHalfSpacedXII 

\usepackage{algorithm}
\usepackage{algpseudocode}
\usepackage{tikz}
\usepackage[table]{xcolor}

\usepackage{natbib}
 \bibpunct[, ]{(}{)}{,}{a}{}{,}%
 \def\bibfont{\small}%
\EquationsNumberedThrough    

\TheoremsNumberedThrough     
\ECRepeatTheorems  %

\MANUSCRIPTNO{IJOC-0001-2024.00}

\begin{document}

\RUNAUTHOR{Lawless et al.}

\RUNTITLE{LLMs for Separator Configuration}

\TITLE{LLMs for Cold-Start Cutting Plane Separator Configuration}

\ARTICLEAUTHORS{%
\AUTHOR{Connor Lawless, Yingxi Li, Anders Wikum, Madeleine Udell, Ellen Vitercik}
\AFF{Management Science and Engineering, Stanford University} 

} 

\ABSTRACT{%
Mixed integer linear programming (MILP) solvers expose hundreds of parameters that have an outsized impact on performance but are difficult to configure for all but expert users. Existing machine learning (ML) approaches require training on thousands of related instances, generalize poorly and can be difficult to integrate into existing solver workflows. We propose a large language model (LLM)-based framework that configures cutting plane separators using problem descriptions and solver-specific separator summaries. To reduce variance in LLM outputs, we introduce an ensembling strategy that clusters and aggregates candidate configurations into a small portfolio of high-performing configurations. Our method requires no custom solver interface, generates configurations in seconds via simple API calls, and requires solving only a small number of instances. Extensive experiments on standard synthetic and real-world MILPs show our approach matches or outperforms state-of-the-art configuration methods with a fraction of the data and computation.
}%





\maketitle

\section{Introduction}

Mixed Integer Linear Programming (MILP) is a remarkably effective mathematical framework for discrete optimization that has been successfully deployed across industries ranging from healthcare \citep{adalkareem2021healthcare} to routing \citep{caceres2014rich} and production planning \citep{pochet2006production}. Modern MILP solvers (e.g. Gurobi and SCIP) expose a number of configurable parameters that control key algorithmic components of the solve---including cutting planes, branching strategies, and the use and scheduling of primal heuristics. By default, MILP solvers use base configurations that are tuned for good average performance over a broad class of potential optimization problems. However, these general-purpose parameter settings can often be improved substantially for specific instances or sets of instances \citep{xu2011hydra,li2024learning}. Motivated by this fact, and in part because configuration for good performance on a particular problem is notoriously difficult even for expert users, a flurry of recent work (see \cite{bengio2021machine} for a comprehensive survey) has focused on tuning parameters for \textit{families} of instances using machine learning (ML).

In this paper, we study the problem of configuring which cutting plane separators to apply when solving a given MILP instance. The integration of cutting planes (or cuts) into MILP solvers has driven impressive 32$\times$ speedups in computation time for branch-and-cut solvers such as CPLEX \citep{achterberg2013mixed} by strengthening linear relaxations at nodes in the branch-and-bound tree. Recent work has shown that better configuration of cutting plane separators alone can achieve up to 72\% speed-ups over SCIP default settings on classic combinatorial optimization problems \citep{li2024learning}. Existing ML-based approaches for cutting plane separator configuration, and IP configuration more broadly, typically involve training an ML model by solving hundreds, if not thousands, of related MILP instances multiple times. 
This effort requires access to both historical data from the same instance family and a hefty computational budget to run all the instances. Furthermore, these methods often employ custom solver interfaces that expose parts of a solver's internal machinery. 
This requirement complicates implementation and restricts use to less powerful, open-source solvers. 
Given the daunting complexity and computational cost of implementing existing ML approaches, in this paper, we study the following research question:

\textit{Can we configure problem-specific cutting plane separators with little to no training data, negligible computation time, and only access to existing solver APIs?} 

We call this problem the \textit{cold-start} cutting plane separator configuration problem. Previous literature has highlighted some potential pathways to an affirmative answer. Bixby and Rothberg remarked that an ideal cutting plane method would \textit{``be able to recognize models to which it can be applied, introducing little or no overhead when the model does not fit the mold"} \citep{bixby2007progress}.
Indeed, a configuration approach that can identify relevant model structure and turn on/off the associated separators could lead to better problem-specific configuration. Furthermore, previous studies on cutting plane separator configuration have validated their models by highlighting that the learned configurations match known results from the literature \citep{li2024learning}, which suggests that existing cutting plane research could provide a good signal for configuration. 

Based on these two observations, large language models (LLMs) offer a promising direction toward generating better problem-specific configurations without training. 
Recent work has highlighted that LLMs are able to identify and model MILP structures from natural language alone \citep{ahmaditeshnizi2024optimus,lawless2024want}. Moreover, LLMs are trained on wide swaths of the internet \citep{achiam2023gpt}, which may include existing work on cutting planes and mathematical programming. However, many challenges arise when using LLMs for configuration. 
First, each MILP solver implements its own (often proprietary) set of cutting plane separators, which can differ from others in name, implementation, or both \citep{achterberg2019gurobi,manual1987ibmcplex}. 
As a result, our task is solver-dependent. Second, LLMs are known to be noisy and to hallucinate (i.e., generate factually incorrect text) \citep{huang2023survey}, which is problematic given that solver performance is extremely sensitive to parameter choices (i.e., adding or removing certain separators). 
More generally, while prior work has shown strong performance for specific separators on some special problem structures, cutting plane selection more broadly is a notoriously difficult problem from both a theoretical and empirical perspective \citep{dey2018theoretical}. 

In this paper, we establish LLMs as a viable tool for performing cold-start algorithm configuration based solely on a natural language description of the problem to be solved. To the best of our knowledge, this is the \textit{first work to explore the use of LLMs for algorithm configuration}. To reduce hallucinations and design solver-specific configurations, we augment base LLMs by providing solver-specific descriptions of cutting plane separators that have been generated by summarizing existing research. Moreover, because LLM outputs are stochastic, we introduce an ensembling technique that first generates a pool of candidate configurations, then clusters them using a $k$-med clustering algorithm to build a small representative set of potential configurations. The final configuration is chosen from among these candidates using either a heuristic in settings with no data, or by performance on a small validation set. We further extend the framework to a text-free setting, motivated by the fact that most solver pipelines operate directly on \texttt{.mps} files without accompanying problem descriptions. 
Across a comprehensive benchmark of MILP instances, we demonstrate that our approach generates high-performing cutting plane configurations that are competitive with existing ML approaches at a fraction of the computational cost.  A preliminary version of this work \citep{lawless2025llms} introduced the baseline LLM-configuration pipeline and provided some initial experiments. This article extends the approach to text-free settings and includes a comprehensive empirical evaluation on large-scale and out-of-distribution MILPs.
\section{Background and Related Work}

A mixed-integer linear program is an optimization problem of the form 
$$
    \argmin_{x} \left\{ c^Tx \mid Ax \leq b, x \in \mathbb{R}^{n-r} \times \mathbb{Z}^r \right\},
$$
where $x$ is a vector of decision variables, $c \in \mathbb{R}^n$ is a vector of objective coefficients, and $A \in \mathbb{R}^{m \times n}$ is a constraint coefficient matrix. The size of a MILP is characterized by the number of constraints ($m$) and variables ($n$) in the problem. 

The standard algorithms for solving MILPs to optimality are variants of the branch-and-bound (B\&B) algorithm \citep{IPref}. At a high level, B\&B recursively partitions the solution space, organizing this partition with a search tree (branch). During search, B\&B uses linear programming (LP) relaxation bounds to prune sub-trees that provably cannot contain an optimal solution (bound). The rate at which sub-trees are pruned depends, in part, on the quality of these LP relaxations. One method that enables higher-quality relaxations is the introduction of cutting planes---valid linear inequalities that tighten the feasible region without excluding any feasible solutions. This augmented approach, referred to as branch-and-cut, can significantly accelerate B\&B by more quickly refining the search space. However, generating cutting planes can be computationally expensive \citep{bixby2007progress}. 

Most MILP solvers implement a diverse set of separators that are used to generate cuts, ranging from general-purpose (e.g., Gomory, MIR, and Cover Cuts) to specialized for specific constraint structures (e.g., knapsack inequalities) \citep{IPref}. Each separator presents a different trade-off between performance improvement and computation time. Typically, separators are run in a prioritized order until a specified number of cuts have been generated. The generated cuts are then added to a cut pool that acts as a buffer to store the cuts and apply the most promising ones to a given B\&B tree node. Well-chosen cutting plane configurations introduce limited overhead by deactivating computationally demanding or ineffective separators, while still improving the quality of the cut pool. The remainder of this section reviews the literature most relevant to our work. 

\subsection{Algorithm configuration}

Our work is closely related to the rich literature on algorithm configuration \citep{balcan2021generalization,balcan2021much}, a field that seeks automatic methods to find the best settings for parameterized algorithms. Early work developed general-purpose configuration methods, with notable frameworks including ParamILS \citep{hutter2009paramils}, SMAC \citep{hutter2011sequential}, and GGA+ \citep{ansotegui2015model}. While these methods are effective in identifying a single configuration with strong performance across a dataset, they can struggle with instance heterogeneity. Instance-specific frameworks, which are customized to individual problem instances using machine learning (ML), were later developed to address this limitation. Prominent examples include ISAC \citep{kadioglu2010isac}, which clusters instances based on features to identify configurations, Hydra \citep{xu2010hydra}, which iteratively constructs portfolios of configurations, and PCIT \citep{liu2019automatic}, which employs advanced ML techniques for automated configuration tuning. 

Early applications of ML to MILP solving emerged from these instance-specific algorithm configuration frameworks. The Hydra-MIP framework \citep{xu2011hydra} specialized Hydra for parameter selection in MILP solvers. At the time, Hydra-MIP demonstrated performance improvements over default settings in the concurrent version of CPLEX, showcasing the potential of automated parameter tuning to improve solver efficiency. However, training these parameter configuration models is computationally intensive. Hydra-MIP is trained over distributions of MILP datasets and, for example, required over 250,000 CPU days of runtime and access to 500 sample MILP instances. Our method distinguishes itself from classic configuration approaches in that it achieves significant performance improvements over solver defaults at little computational cost, and that it uses a novel input modality---natural language descriptions---to develop the configurations.

\subsection{Machine learning for combinatorial optimization}

The existing body of research on ML for combinatorial optimization (CO) can be broadly categorized into two approaches: (1) integrating ML models into existing MILP solvers to enhance their performance, and (2) training ML models to directly make decisions that constitute part or all of the solution \citep{khalil2017learning,Deudon2018LearningHF,Kool2018AttentionLT,ye2024deepaco,jiang2024ensemble}. Our work aligns closely with the first category. 

In recent years, for example, there has been an explosion of work on using ML to tune various search heuristics within the B\&B algorithm~\citep{khalil2022mip,gasse2019exact,khalil2016learning} to accelerate MILP solutions. The use of ML for cutting plane selection has also garnered significant attention, with ML guiding cut generation \citep{deza2024learn2aggregate,guaje2024machine,Balcan21:Sample,Balcan22:Structural}, selecting cuts from a generated pool \citep{tang2020reinforcement,paulus2022learning}, or configuring separators that generate high-quality pools of cuts \citep{li2024learning}. Among these, the work by \citep{li2024learning} is particularly relevant to ours, proposing a neural contextual bandit model that predicts solver-specific cutting plane configurations for various problem distributions. To achieve strong performance, this model requires solving thousands of MILPs from the same distribution during model training, as well as a custom SCIP solver interface to change the configuration at different depths of the B\&B tree. 

In contrast to the work of \citet{li2024learning}, we focus on the problem of selecting an \textit{instance-agnostic configuration} (i.e., one configuration to use for all instances coming from a given problem distribution). We do so for two practical reasons. First, learning instance-specific configurations often requires training ML models to select a configuration from a portfolio of potential configurations \citep{xu2011hydra,li2024learning} based on high-dimensional instance-specific features (e.g., a bipartite graph representation of the problem). These approaches also require a large computational budget to train the ML model, requiring thousands of MILP solves. In resource-constrained settings, there are often not enough instances to effectively train such models, nor sufficient computational resources to support the training process. Second, deep learning-based configuration algorithms for instance-specific inference often require additional computational resources and memory, which makes them impractical to deploy in some settings. Our work also differs from \citep{li2024learning} in that we configure separators once rather than at each level of the  B\&B tree, as this approach does not require any custom solver interface. 

\subsection{LLMs and optimization}

Recent advances in large language models (LLMs) have led to impressive performance in tasks across a number of domains \citep{chen2023teaching,yuan2022wordcraft,zhang2023prompting}. However, LLMs are known to hallucinate, generating seemingly plausible but factually incorrect text (see \citep{zhang2023siren} for a comprehensive survey). They are also sensitive to prompts: semantically similar prompts can result in significantly different outcomes \citep{kojima2022large}. Recent literature focuses on addressing this problem through prompt optimization \citep{lester2021power,li2021prefix,pryzant2023automatic} and augmentation of LLMs using interactive tools \citep{nakano2021webgpt}. 

Despite these shortcomings, the ability of LLMs to leverage natural language points us to new possibilities for optimization. Recent work has explored LLMs for optimization modeling \citep{ramamonjison2022augmenting,tsouros2023holy,ahmaditeshnizi2023optimus,ahmaditeshnizi2024optimus,xiao2023chain,astorga2024autoformulation,tang2024orlmtraininglargelanguage}. Building on these capabilities, LLMs have powered chatbots that allow users to interact with an underlying optimization model in the context of supply chain management \citep{li2023large}, meeting scheduling \citep{lawless2024want}, debugging infeasible models \citep{chen2023diagnosing}, and evaluating the equivalence of different formulations \citep{zhaiequivamap}. Our work differs in that we use LLMs to improve the efficiency of solving an optimization problem described in natural language rather than formulating the MILP. Most similar to our work, a line of research has investigated the use of LLMs to tune hyperparameters \citep{mahammadli2024sequential,zhang2023using,custode2024investigation}. These approaches employ LLMs as \emph{optimizers} to iteratively refine hyperparameter selection and require repeatedly solving MILP instances. In contrast, our work focuses on the cold-start problem where very few MILP instances need to be solved, leverages existing domain knowledge in optimization to improve LLM performance on this task, and introduces novel strategies to ensemble noisy LLM outputs. 

\section{LLMs for Separator Configuration}
In this paper, we study a variant of the cutting plane separator configuration problem introduced in \citet{li2024learning}. The input to the problem consists of a MILP solver with $M$ different separator algorithms (which take as input a MILP and return valid cutting planes) and $S$ possible settings for each separator. For example, Gurobi supports $M = 21$ different separator algorithms, each with $S = 3$ possible settings (disable, use, and use aggressively), for a total of $|S|^M$ candidate configurations. The goal of the separator configuration problem is to find the best setting of the separators for a given MILP. 

Our default method assumes access to a natural language representation of the MILP that includes a short text description and \LaTeX~formulation for the problem. If ${\cal X}$ is the unknown distribution of MILP instances corresponding to the natural language description, we aim to select a high-performing \textit{instance-agnostic configuration} $\mathcal{C}$ (i.e., one configuration to use for all instances coming from ${\cal X}$), given only a small validation dataset $\mathcal{K}_\textup{val}$ sampled from ${\cal X}$. We refer to the setting where ${\cal K}_\textup{val} = \emptyset$ as the \textit{cold-start configuration problem}. The performance of a separator configuration for a particular solver can be measured by a number of metrics. Following the setup of \citet{li2024learning}, we use relative improvement in solve time over the default solver configuration---a detailed description of this metric can be found in our experimental setup in Section \ref{subsec: setup}.

\begin{figure}[!t]
    \centering
    \centerline{
    \includegraphics[width=\textwidth]{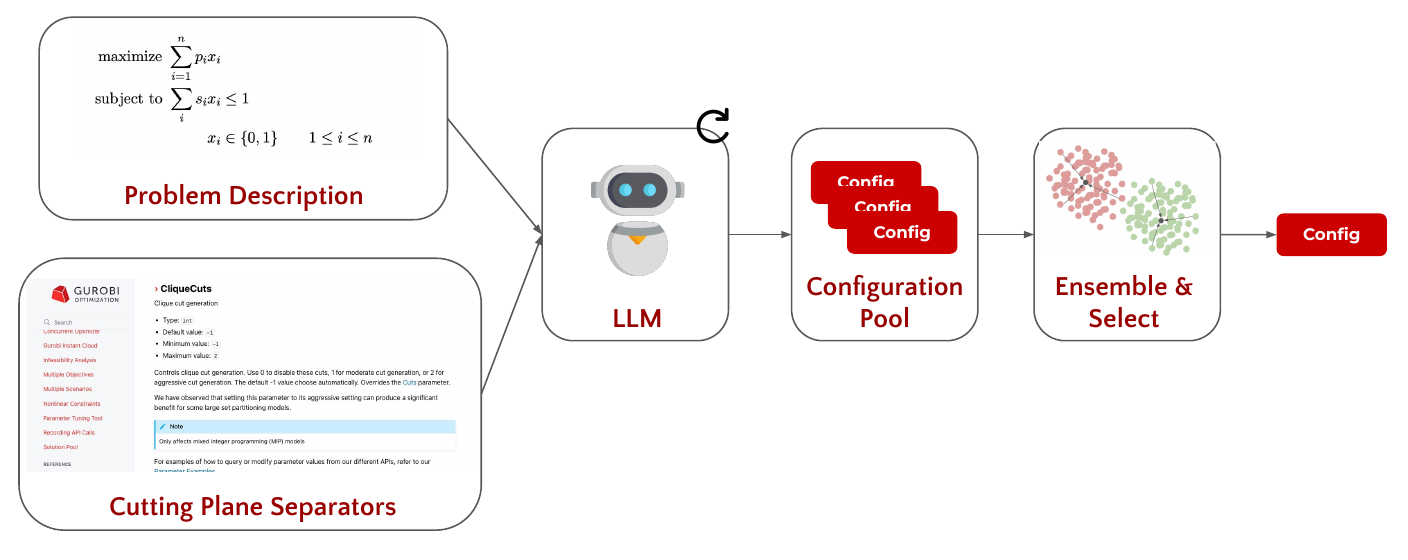}
    }
    \caption{Overview of cutting plane separator configuration algorithm.}
    \label{fig:algo}
\end{figure}

Our approach uses LLMs to generate promising candidate separator configurations, exploiting both problem-specific context and prior knowledge encoded in the LLM. Figure \ref{fig:algo} summarizes our separator configuration procedure. In Section \ref{subsec:generation}, we detail how we leverage LLMs to generate candidate separator configurations based on natural language descriptions of the optimization problem and available cutting plane separators. In Section \ref{subsec:ensembling}, we show how we ensemble a pool of candidate configurations into a single final configuration via k-medoids clustering and a validation step on a small dataset of instances. Finally, in \ref{subsec:text_free} we extend our method to the more challenging \emph{text-free} setting in which only a \texttt{.mps} file is provided.

\subsection{Generating configurations} \label{subsec:generation}

\begin{figure}[!t]
    \centering
    \centerline{
    \includegraphics[width=\textwidth]{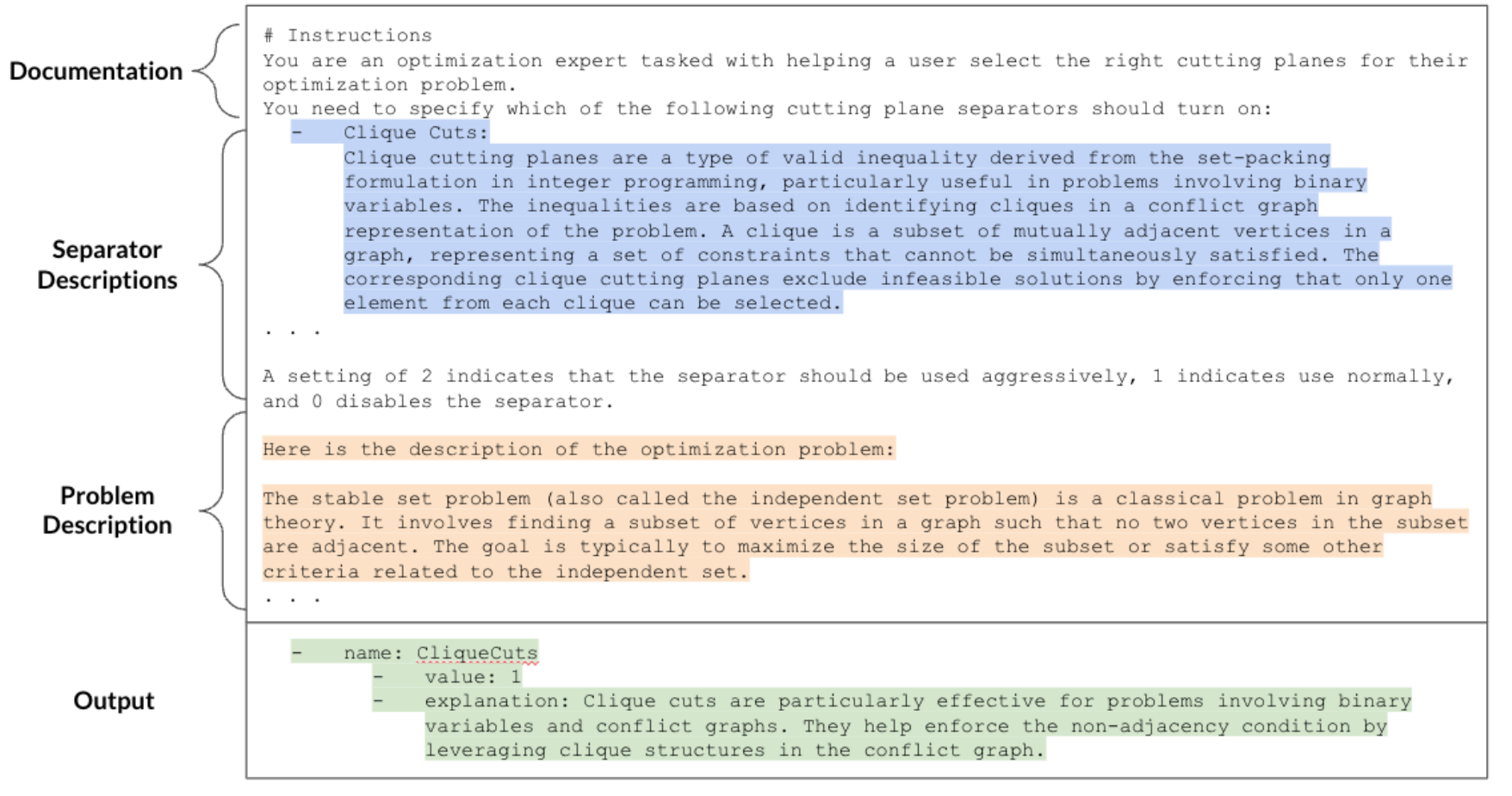}
    }
    \caption{Sample prompt for generating a separator configuration. Blue (orange) lines indicate solver-specific (problem-specific) information in the prompt. Green lines indicate a sample LLM output.}
    \label{fig:prompt}
\end{figure}

To generate high-performing separator configurations from LLMs, we construct a prompt that includes both problem-specific and solver-specific information (see Figure \ref{fig:prompt} for a sample prompt). 
The prompt includes a text description of the optimization problem as well as its corresponding \LaTeX~formulation. For the models evaluated in this paper, this natural language description was generated using an LLM to summarize either a relevant textbook section or research paper that introduced the model, and then checked for accuracy by an optimization expert. Although a natural language description is an additional requirement compared to existing algorithm configuration approaches, which typically rely on access to the model directly as a \texttt{.mps} or \texttt{.lp} file, the optimization user often already has a text description of the problem, or can generate one easily, in most practical applications. In Section \ref{subsec: ablations}, we show that our approach is robust to the type of information included about the model. 

Each MILP solver has a different set of separators available, each with its own solver-specific parameter names, and so we limit LLM hallucinations by including a solver-specific description of available cutting plane separators in the prompt. These descriptions include the specific parameter name as well as a natural language description of the separators. To generate these descriptions, we use an LLM to summarize existing research papers that introduce the different separators, producing concise descriptions that specify the types of problem structure the cutting planes can be used on (e.g., \textit{Implied Bound Cuts can be used in settings where there is a logical implication between binary and continuous variables}). We further augment these sources with online resources, such as slides and survey papers, and expert knowledge from the integer programming academic community. 
An inherent limitation of our approach is that for closed-source solvers (e.g., Gurobi), it is impossible to know exactly what implementation and separator procedure is used for a given family of cutting planes. Consequently, these descriptions offer only a broad summary. 
Nonetheless, these initial descriptions perform well in practice and can only be improved by richer information about the underlying separator algorithms. 
For open-source solvers, additional information, such as code implementations, is publicly available and thus can provide even more context to an LLM-generated summary of the separators. While we do not exploit this information in our descriptions, effectively leveraging this additional context, as well as other information sources more broadly, is an important direction for future research.

\subsection{Ensembling configurations} \label{subsec:ensembling}

\begin{figure}[!t]
    \centering
    \centerline{
    \includegraphics[width=\textwidth]{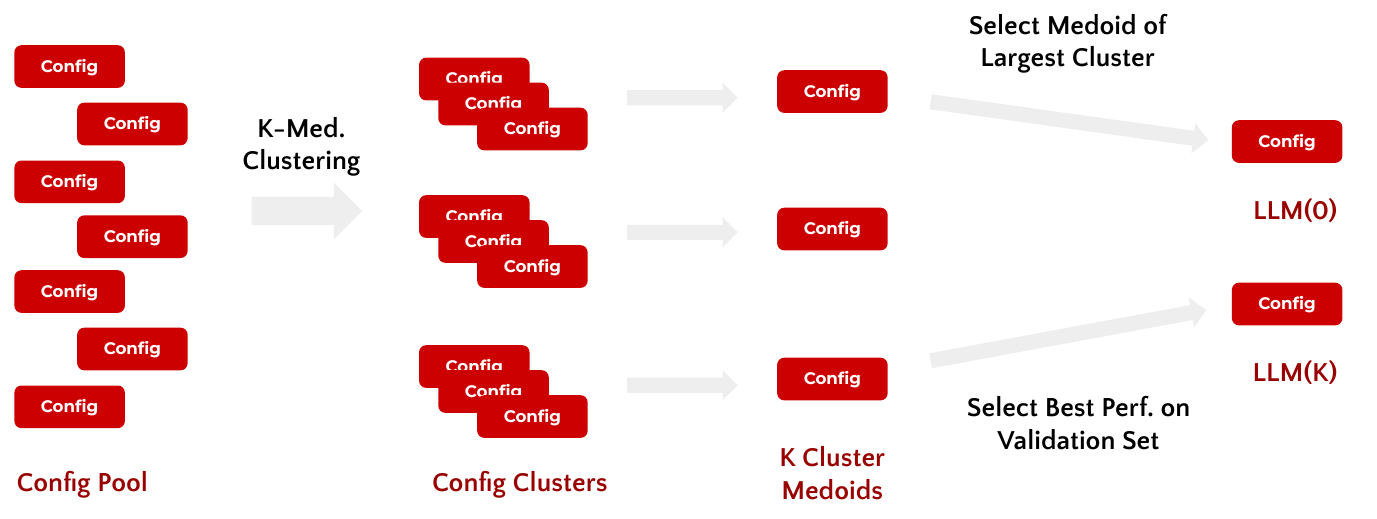}
    }
    \caption{Overview of pipeline for ensembling and selecting from the configuration pool. The pool of configurations is first clustered into $k$ clusters using the $k$-medoids algorithm. The final configuration is then either (a) selected as the medoid of the largest cluster for LLM(0), or (b) selected by testing the $k$ cluster medoids on the validation dataset and choosing the one with the best performance.}
    \label{fig:ensemble}
\end{figure}

LLMs have stochastic outputs: re-running the same prompt multiple times can produce different outcomes. Motivated by similar approaches \citep{cobbe2021training,bansal2024smaller}, we exploit this feature by sampling $M$ candidate separator configurations using the LLM and then \textit{ensemble} them together into one final configuration. Initial experiments showed that separators used in LLM-generated configurations were correlated. Hence, we cluster the pool of configurations into $k$ clusters using a $k$-medoids clustering algorithm to find structurally distinct configurations, where $k$ represents a hyperparameter in our approach.
We use $k$-medoids instead of $k$-means to ensure that the selected configuration is a configuration that was output by the LLM. To select a final configuration from the $k$ medoids of the generated clusters, we evaluate each medoid on the problems in $\mathcal{K}_\textup{val}$, then select the configuration with the largest median improvement over the solver's default configuration. Note that the number of clusters $k$ can, and should, be chosen to match the computational budget of the application. Larger values of $k$ correspond to trying more candidate configurations and are expected to increase the performance of the algorithm on the validation dataset. In the cold-start setting, without any related MILP instances, we select the medoid of the largest cluster as a heuristic. 
\subsection{Text-free configuration} \label{sec:text_free_meth}
A limitation of our standard configuration pipeline is its reliance on a natural language description of the optimization problem, which is often unavailable in solver pipelines that operate solely on standard input formats such as \texttt{.mps} files. To address this, we introduce a text-free variant that generates plausible problem descriptions directly from the \texttt{.mps} file. Specifically, we process the \texttt{.mps} file to extract the distribution of constraint types present in the instance, following the categorization scheme from MIPLIB 2017 \citep{gleixner2021miplib}. Constraint types are a natural feature for the separator configuration problem, as many separators are designed to exploit specific constraint structures (e.g., network flow constraints). Given a list of constraint types present in the problem, we then prompt an LLM to generate candidate natural language descriptions consistent with the identified constraint types. Each candidate description is passed through the same configuration generation pipeline outlined in Section~\ref{subsec:generation}, and the resulting configurations are added to the configuration pool before applying the ensembling strategy of Section~\ref{subsec:ensembling}. An additional benefit of this approach is that it enables instance-specific configuration: rather than relying on a general description of a problem family, we can produce separator configurations tailored to a single MILP instance.
\begin{figure}[!t]
    \centering
    \centerline{
    \includegraphics[width=\textwidth]{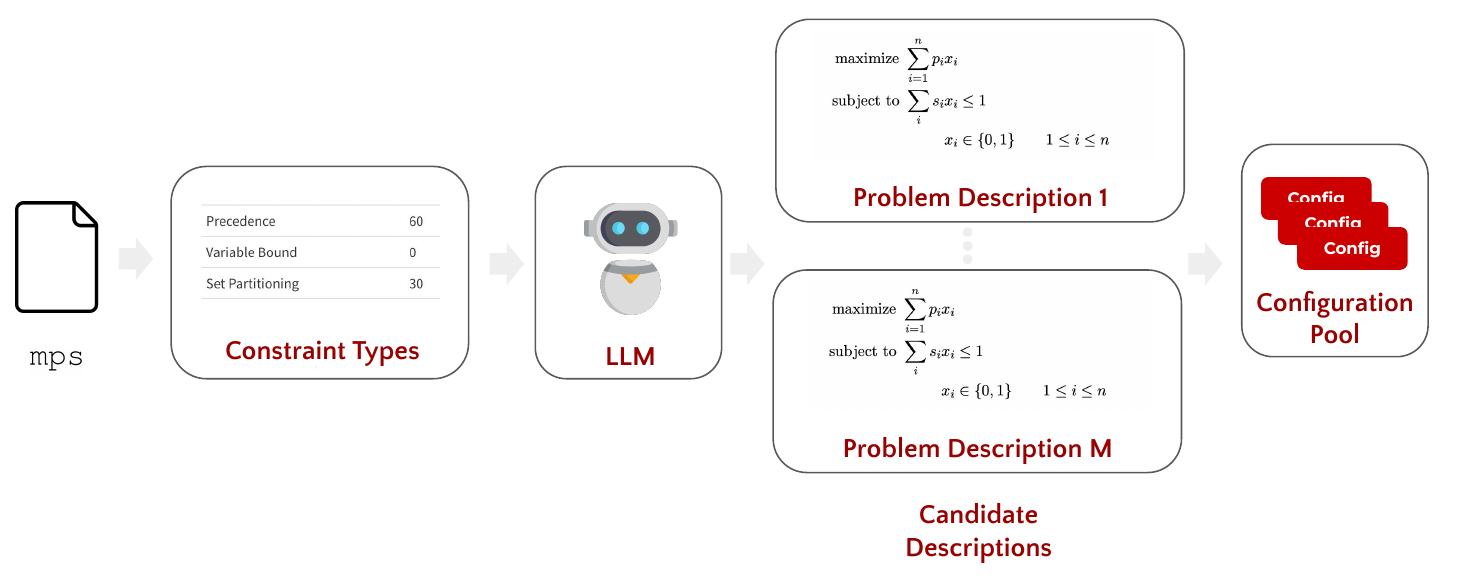}
    }
    \caption{Overview of pipeline for generating configuration pool without any text description.}
    \label{fig:textfree}
\end{figure}

\section{Experiments}

To highlight the practical utility of our approach for solver configuration, we evaluate our approach across a comprehensive suite of classic MILP benchmarks, real-world industrial problems, and completely novel MILP instance families. We evaluate two variants of our framework: \textsc{LLM}($k$), our default pipeline which ensembles configurations generated from problem descriptions, and LLM-no-text($k$), which operates directly on \texttt{.mps} files without textual input. Here $k$ denotes the number of clusters generated during the ensembling process described in Section \ref{subsec:ensembling}. For our experiments, we generate 100 configurations to form the configuration pool, and use GPT-4o \citep{hurst2024gpt} as the underlying LLM. In the interest of brevity, we highlight results for configuring SCIP in the main body of the paper and include results on Gurobi in Appendix \ref{app:gurobi}. All code for our experiments is available at \url{https://github.com/OptiMUS-optimization-modeling/LLM-Solver-Configuration}

The remainder of this section is organized as follows. Section \ref{subsec: setup} introduces the experimental setup, including datasets, evaluation metrics, baselines, and hardware. Section \ref{subsec: performance} benchmarks LLM($k$) against baselines on standard MILP families, while Section \ref{subsec:ood} tests its ability to generalize to large-scale and out-of-distribution problems. Section \ref{subsec:text_free} evaluates the text-free pipeline LLM-no-text($k$), and Section \ref{subsec: ablations} presents ablation studies that isolate the contributions of individual components of our approach.


\subsection{Setup} \label{subsec: setup}

\paragraph{\textbf{\textup{Data.}}} We evaluate our methods on a number of  MILP datasets. These include standard benchmarks from \citet{tang2020reinforcement} and Ecole \citep{gasse2019exact}, real-world MILP distributions from the Distributional MIPLIB library \citep{huang2024dmiplib}, and novel ``out-of-distribution'' instances from the MILP-evolve dataset \citep{li2024foundation}. Table \ref{table: data-description} provides an overview of the standard distributional MILP families used in our base experiments. 
To test generalization, we test on both larger instances from the same families and completely novel composite formulations from the MILP-evolve dataset, which are generated to mimic real-world optimization scenarios. For each distributional MILP family considered, we maintain a small validation set $\mathcal{K}_\textup{val}$ and a larger evaluation set $\mathcal{K}_\textup{eval}$ of 30 and 100 MILP instances, respectively. We also evaluate our text-free pipeline on a subset of instances from MIPLIB 2017. Additional details on each of the three datasets are included in Appendix \ref{app:exp-details}.

\begin{table}[t] 
\centering
    \caption{Families of standard benchmark MILPs considered}
    \label{table: data-description}
    \begin{adjustbox}{max width=\textwidth,center}
\begin{tabular}{
    @{} l
    @{\hspace*{8mm}}     c
    @{\hspace*{8mm}}     c
    @{\hspace*{8mm}}     c
    @{\hspace*{8mm}}     c
    @{\hspace*{8mm}}     c
}
\toprule
\textbf{Name}  & \textbf{\# vars} & \textbf{\# constrs} & \textbf{SCIP solve time (s)} &  \textbf{Source} \\
\midrule
Binary Packing & 300 & 300 & 1.76 &   \citep{tang2020reinforcement}  \\\addlinespace[1mm]
Capacitated Facility Location & 100 & 100 &  20.11  & \citep{gasse2019exact} \\\addlinespace[1mm]
Combinatorial Auction & 100 & 500 & 2.18 & \citep{gasse2019exact}  \\\addlinespace[1mm]
Maximum Independent Set & 500 & 1088 & 1.916 & \citep{gasse2019exact}   \\\addlinespace[1mm]
Max Cut & 54 & 134 &  0.49 & \citep{tang2020reinforcement}   \\\addlinespace[1mm]
Packing & 60 & 60 &  12.25 & \citep{tang2020reinforcement}    \\\addlinespace[1mm]
Set Cover & 500 & 250 & 1.26 & \citep{gasse2019exact}   \\\addlinespace[1mm]
\midrule \addlinespace[2mm]
Load Balancing & 64340 & 61000 & 14.68$^1$  & \citep{gasse2022machine}   \\\addlinespace[1mm]
Middle-Mile Consolidation\\ Network Design (MM) & 569 & 248 & 1.33 & \citep{greening2023lead}  \\\addlinespace[1mm]
\bottomrule
\end{tabular}
\end{adjustbox}

\scriptsize{$^1$ Solved to 10\% optimality gap \hfill}
\end{table}

\paragraph{\textbf{\textup{Evaluation.}}} Following the setup of \citet{li2024learning}, we measure the performance of a separator configuration for a particular solver by its relative solve time improvement over that solver's default configuration. Explicitly, the (mean) relative solve time improvement ($\delta$) of a separator configuration $\mathcal{C}$ on a MILP instance $x \in {\cal K}$ is
\[\delta(x, \mathcal{C})= 100 \cdot \frac{t^\mathcal{C}_x - t_x}{t_x},\]
where $t_x$ and $t^{\mathcal{C}}_x$ are the mean solve time of instance $x$ under the default separator configuration and $\mathcal{C}$, respectively. Empirically, we estimate these means by taking the average solve time across 10 MILP solves for each configuration, under different random seeds. We solve most problem families to optimality, excluding the Load Balancing and MIPLIB 2017 instances, which are solved to a 10\% optimality gap (as done in \citet{li2024learning}).

We impose no time limits when solving using default solver configurations, but LLM-generated configurations are limited to $2.5\times$ the default solve times (either during validation to select a configuration, or during evaluation). These time limits censor our reported relative solve time improvement by capping relative performance degradation at -150\% per instance. However, we report medians which are largely insensitive to this kind of censoring. Finally, to mitigate the effects of increased CPU times from parallel processing, the solve times we report for SCIP are wall times (s), while solve times for Gurobi are measured in work units. A work unit corresponds roughly to one second, but it is a deterministic unit of measurement for the same instance and hardware, regardless of parallelization.

\paragraph{\textbf{\textup{Baselines.}}}  We consider baselines and methods for configuration proposed in \citet{li2024learning}. For each MILP family, we include results for:
 (1) \textsc{Pruning} turns off any cutting planes that were not used while solving the validation instances $\mathcal{K}_\textup{val}$ with default settings. (2) \textsc{Search}$(d)$, the instance-agnostic configuration method from \citet{li2024learning}, samples $d$ candidate configurations uniformly at random, then applies the one with the best median performance on the validation set $\mathcal{K}_\textup{val}$. We refer to $d$ as the depth of the search. Note that we implicitly also compare to default separator settings in Gurobi and SCIP, as our performance metric measures the relative time improvement over default.

\paragraph{\textbf{\textup{Hardware.}}}  All MILPs are solved on a distributed compute cluster whose nodes are each equipped with two 64-core AMD EPYC 7763 CPUs, for a total of 256 threads. We fully parallelize when evaluating Gurobi separator configurations, and impose a limit of 4 threads per process when evaluating configurations for SCIP. This restriction is standard in the ML for integer programming community (see, e.g. \citet{gupta2022lookback}) to limit the impact of thread competition on solve time.

\subsection{Performance on standard distributional MILP benchmarks} \label{subsec: performance}

\begin{table}[!t] 
\centering
    \caption{Median (higher is better) and IQR of relative solve time improvements (\%) over default SCIP separator configuration.}
    \label{table: scip-benchmark}
    \begin{adjustbox}{max width=\textwidth,center}
\begin{tabular}{
    @{} l
    @{\hspace*{12mm}}     c
    @{\hspace*{8mm}}     c
    @{\hspace*{8mm}}     c
    @{\hspace*{8mm}}     c
    @{\hspace*{8mm}}     c
                         c
    @{\hspace*{4mm}}     c
}
\toprule
 & \multicolumn{5}{c}{\textbf{SCIP configuration}} \\
 \cmidrule(lr){2-6}
\textbf{Problem} & \textsc{Pruning} & \textsc{Search}(5) &  \textsc{Search}(500) & \textsc{LLM}(0) & \textsc{LLM}(5)\\
\midrule
 \textbf{Bin. Pack.} & 1.33 (7.78) & 9.23 (28.37) & 39.3 (27.72) & 16.76 (23.79) & 38.35 (27.38) & \\\addlinespace[1mm]
 \textbf{Cap. Fac.} & -0.64 (14.68) & 9.57 (24.42) & 2.72 (3.09) & 7.61 (17.19) & 26.12 (30.88) \\\addlinespace[1mm]
\textbf{Comb. Auc.} & 1.96 (14.89) & 58.1 (29.69) & 64.01 (49.66) & 21.06 (31.67) & 63.59 (49.74) \\\addlinespace[1mm]
\textbf{Ind. Set} & 2.07 (22.42) & 26.95 (46.08) & 67.01 (22.7) & 21.6 (49.37) & 71.95 (21.95) \\\addlinespace[1mm]
\textbf{Max. Cut} & -2.18 (5.17) & 17.72 (29.47) & 69.63 (12.43) & 71.43 (11.8) & 71.01 (11.73) \\\addlinespace[1mm]
\textbf{Pack.} & 15.87 (47.68) & -13.81 (71.34) & 24.49 (42.57) & 15.09 (40.53) & 25.51 (42.27) \\\addlinespace[1mm]
\textbf{Set Cov.} & 6.62 (13.6) & -10.04 (37.5) & 61.08 (22.56) & 61.72 (21.96) & 61.74 (22.39) \\\addlinespace[1mm]
\midrule
\textbf{Load Bal.} & 0.08 (2.51) & -150.01 (0.01) & -50.02 (0.03) & 0.0 (23.43) & 6.37 (13.38) \\\addlinespace[1mm]
\textbf{MM} & -0.12 (6.05) & -8.83 (91.58) & 50.03 (45.82) & -6.52 (43.54) & 53.3 (47.42) \\\addlinespace[1mm]
\bottomrule
\end{tabular}
\end{adjustbox}
\end{table}

\begin{figure}[!t]
    \centering
        \begin{subfigure}
      \centering
      \includegraphics[width=0.45\textwidth]{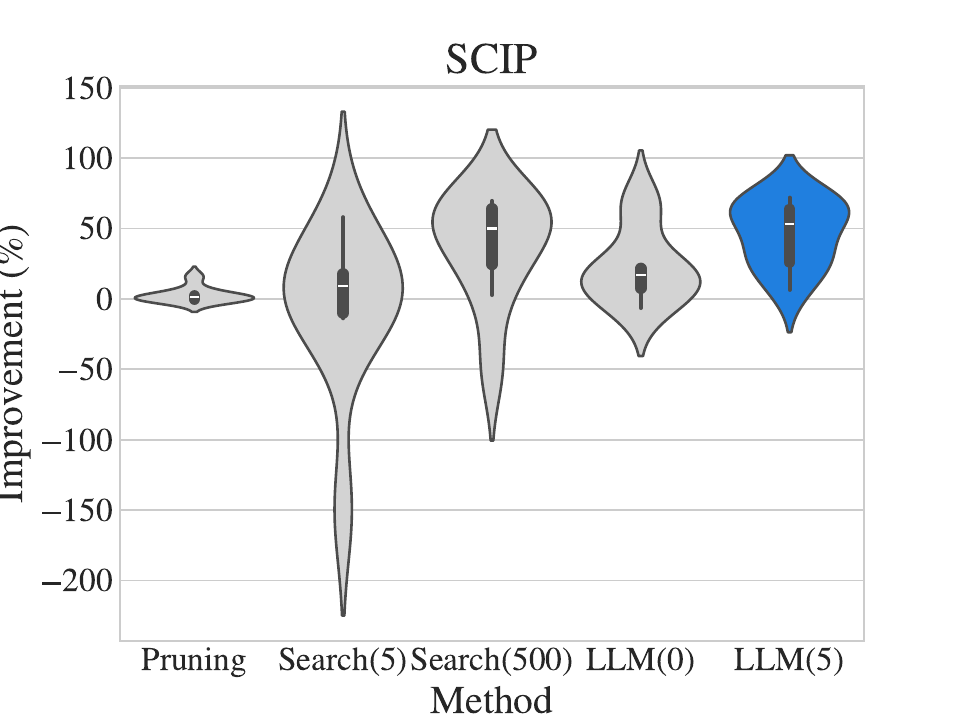}
      \includegraphics[width=0.45\textwidth]{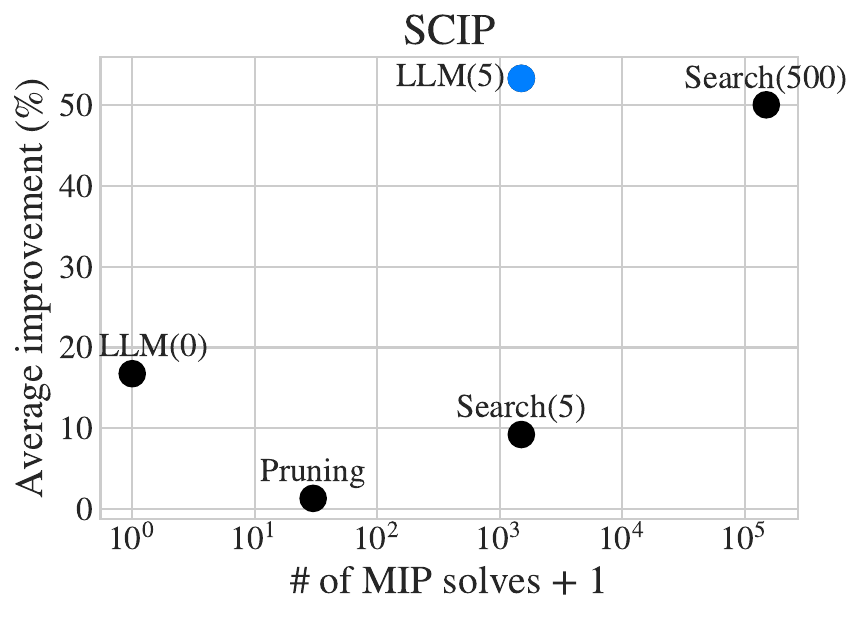}
    \end{subfigure}
    \caption{Distribution of median performance for standard benchmark datasets (higher is better) relative to SCIP default. Note that \textsc{LLM}(5) is able to achieve comparable performance to Search(500) at 1\% of the computational effort.}
    \label{fig: violin_plot}
\end{figure}

Table \ref{table: scip-benchmark} reports solve time improvements for separator configurations generated by our LLM-based algorithm and baselines compared to SCIP defaults. For each method, we report the median and interquartile range of the improvements $\{\delta(x, \mathcal{C}): x \in \mathcal{K}_\textup{eval}\}$ over the evaluation set $\mathcal{K}_\textup{eval}$. Large IQRs reflect the heterogeneity of MILP instances in our evaluation datasets, which has been observed in prior work \citep{li2024learning,wang2023learning}. A horizontal line divides MILPs whose generators come from standard benchmarks vs.~real-world MILPs. Figure \ref{fig: violin_plot} summarizes the results of Tables \ref{table: scip-benchmark} and \ref{table: gurobi-benchmark}. It shows the distribution of improvements 
and highlights the trade-off between the number of MILPs solved and performance for each configuration method.

\begin{figure}[!t]
    \centering
    \centerline{
    \includegraphics[width=0.5\textwidth]{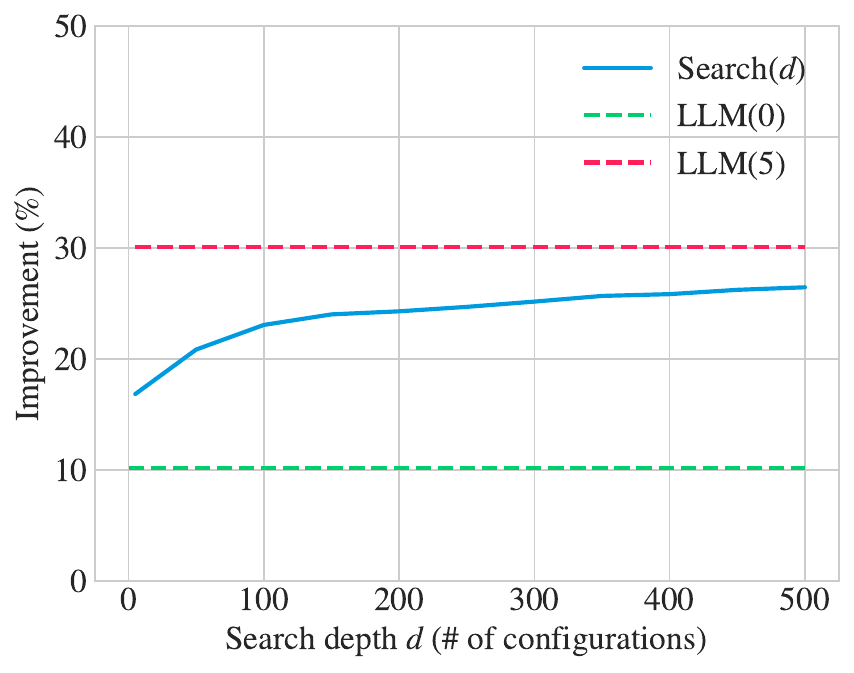}
    }
    \caption{Median performance gain of \textsc{Search}($d$) on Max Cut instances relative to SCIP default, as a function of the number of candidate configurations. Computing \textsc{Search}($d$) requires solving $300 d$ MILPs, while LLM(0) and LLM(5) solve 0 and 1,500 MILPs, respectively.}
    \label{fig: search-depth}
\end{figure}

\textsc{LLM}(5) improves runtime significantly (6-71\%) over the SCIP default configuration. Moreover, LLM-based separator configuration performs well both for problem families that are well-documented (e.g., independent set) and also for non-standard MILP formulations (e.g., Load Bal.), which are likely to be out-of-distribution. In the cold-start setting, our approach \textsc{LLM}(0), which does not solve any MILP instance, outperforms SCIP default settings in all but one problem family. Compared to existing methods, we consistently outperform the \textsc{Pruning} heuristic and are often able to match the improvement of the computationally intensive \textsc{Search}(500) procedure at a fraction of the cost: while LLM(5) can be computed by testing only 5 configurations on ${\cal K}_{val}$, $\textsc{Search}(500)$ requires trying 500 configurations for a total of 150,000 MILP solves. Figure \ref{fig: violin_plot} shows that our approach yields a new point on the Pareto frontier trading off computation time vs performance. At the same computational budget of testing five configurations, our approach \textsc{LLM}(5) outperforms \textsc{Search}(5) on all problem families. Figure \ref{fig: search-depth} shows the trade-off between performance and computational cost for \textsc{Search}($d$) in terms of search depth for one problem family. Lowering the search depth reduces the number of MILP solves needed but also degrades performance.

\subsection{Performance on out-of-distribution instances} \label{subsec:ood}

Real-world MILPs often involve complex formulations that are not covered by existing literature or canonical datasets. As such, a central question is the extent to which the performance gains of LLM-based methods for solver configuration can generalize to unseen formulations that are out-of-distribution for the LLM. We test generalization along two dimensions: unseen problem families and increasing formulation size. 

\begin{figure}[!tb]
    \centering
    \includegraphics[width=0.7\linewidth]{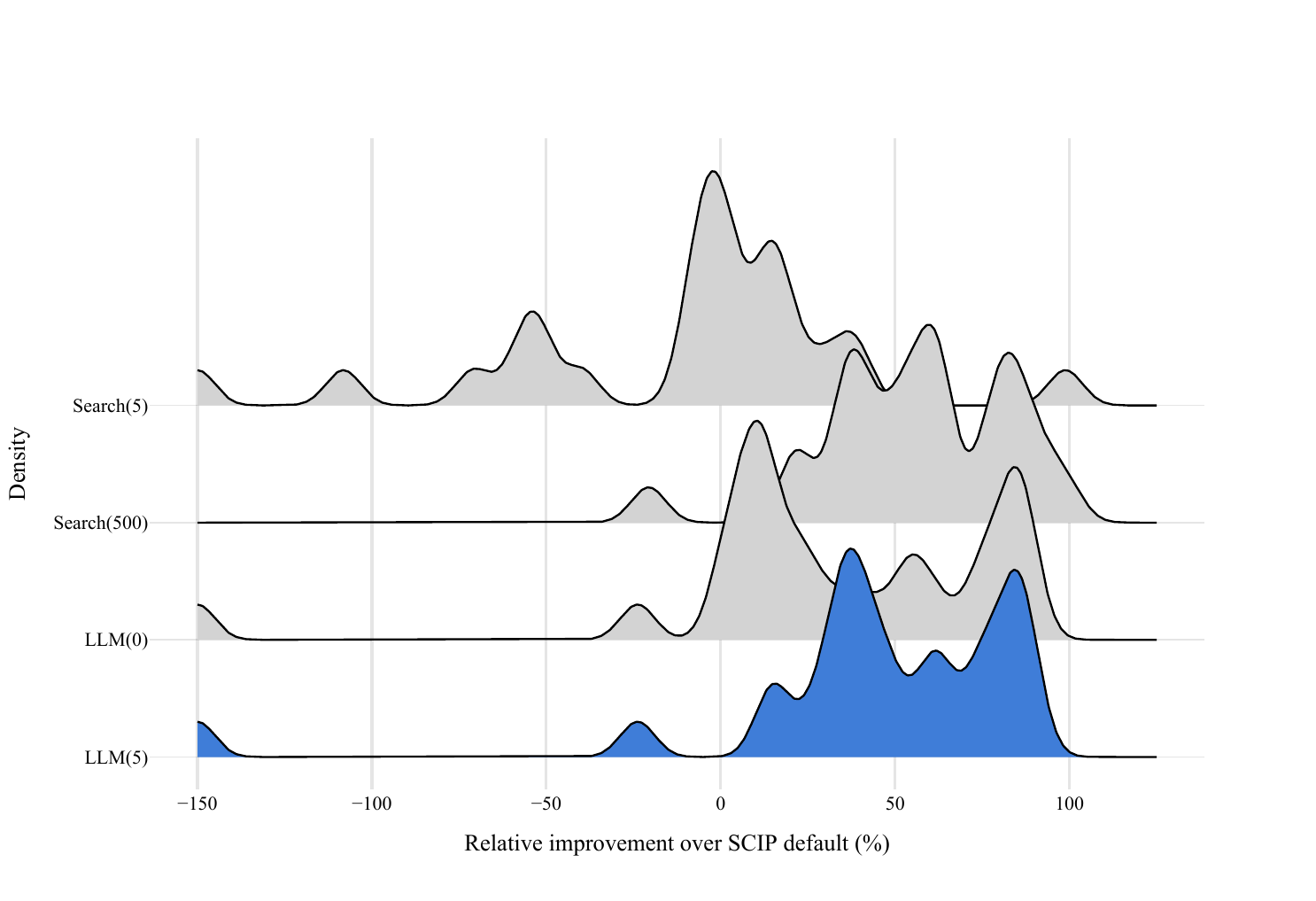}
    \caption{Distributions of median solve time improvements for all easy foundational MILP problems with a text description (SCIP). See  Tables~\ref{tab:foundation_scip} and \ref{tab:foundational_gurobi} in the appendix for more details.}
    \label{fig:foundation-miplib}
\end{figure}

For the former, we benchmark our method's performance on the MILP-evolve dataset introduced by \citet{li2024foundation}. This dataset was designed to be highly diverse to mimic real-world optimization scenarios. We consider all MILP-evolve problems with text descriptions classified as ``easy''. Figure~\ref{fig:foundation-miplib} summarizes solve time improvements achieved by separator configurations generated by our LLM-based approach and by baseline methods, relative to SCIP defaults (detailed results for SCIP and Gurobi are provided in Tables~\ref{tab:foundation_scip} and \ref{tab:foundational_gurobi} in the Appendix). 
Without solving a single MILP, \textsc{LLM}(0) achieves improvements of $5 - 89\%$ across 23 of the 25 MILP families compared to SCIP default, and $0 - 28\%$ across 21 families compared to Gurobi defaults. Under the same computational budget, \textsc{LLM}(5) outperforms the computationally intensive \textsc{Search}(5) on 22 of 25 families by margins of $1 - 142\%$, and either matches (within $5\%$) or surpasses \textsc{Search}(500) on 14 families—while using only $1\%$ of the computational budget.

 To evaluate size generalization, we extend our results for standard MILP benchmarks from Section~\ref{subsec: performance} to larger, more computationally challenging instances from the same problem classes. For each problem variant (family and size), \textsc{LLM(5)} again runs validation to select from among the 5 candidate configurations identified by our aggregation procedure before evaluating on 3 runs of 100 test instances. Tables~\ref{table:scip-size} and \ref{table:gurobi-size} show performance scaling over problems of increasing difficulty for each MIPLIB family we consider---we report the median runtime improvement (\%) over solver defaults for instances solved by both \textsc{LLM(5)} and the solver default, and the median difference in MIP gap (\%, larger is better) for instances that neither method solves to optimality. Overall performance depends heavily on problem class, with improvements generally diminishing as problem difficulty increases. This trend reflects the fact that for complex MILP formulations, solve time can be highly sensitive to over-generation of cutting planes, or missing key separators that enable faster fathoming in branch-and-bound. For example, \textsc{LLM(5)} generally exhibits strong performance for set covering, but misses a network-based cutting plane family for consolidation network design that is critical to efficiently solving at a large scale.

\begin{table}[!t]
\centering
\caption{Performance scaling across MIPLIB families of increasing size and difficulty (SCIP).}
\label{table:scip-size}
\begin{adjustbox}{max width=\textwidth,center}
\begin{tabular}{llccccc}
\toprule
 & \multicolumn{5}{c}{\;\;\;\textbf{Performance metrics}} \\
\cmidrule(lr){3-6}
\textbf{Problem} & \textbf{Difficulty} &\;\; MIP gap difference (\%)$\uparrow$ \;\;& Improvement (\%)$\uparrow$ \;\; & LLM(5) \# solved  & Default \# solved \\
\midrule
\midrule
\multirow{3}{*}{\shortstack[l]{\textit{Combinatorial}\\\textit{auction}}} 
& very easy & - & 64.4 (24.0) & 300/300 & 300/300 \\ 
& medium    & - & 8.40 (26.8) & 300/300 & 300/300 \\
& very hard & 0.17 (0.8) & -  & 0/300   & 0/300   \\
\midrule
\multirow{2}{*}{\shortstack[l]{\textit{Capacitated facility}\\\textit{location}}} 
& easy & - & 23.38 (35.2)   & 300/300 & 300/300 \\
& medium & 0.14 (0.0) & 5.18 (30.1)  & 297/300 & 297/300 \\
\midrule
\multirow{1}{*}{\shortstack[l]{\textit{Load balancing}}}
& hard          & -0.38 (0.7) & -       & 0/300   & 0/300   \\
\midrule
\multirow{3}{*}{\shortstack[l]{\textit{Maximum}\\\textit{independent set}}}
& easy & - & 80.79 (9.7)   & 300/300 & 300/300 \\
& medium & 1.69 (0.3) & 81.10 (14.3) & 291/300 & 210/300 \\
& very hard & 0.23 (2.1) & - & 0/300 & 0/300   \\
\midrule
\multirow{3}{*}{\shortstack[l]{\textit{Middle-mile}\\\textit{consolidation}\\\textit{network design}}}
& very easy  & -      & 51.50 (71.8)   & 300/300 & 300/300 \\
& medium  & -1.83 (0.0) & -15.15 (97.5)& 288/300 & 297/300 \\
& hard & -2.54 (1.6)& -146.53 (0.0)& 3/300   & 33/300  \\
\midrule
\multirow{3}{*}{\shortstack[l]{\textit{Set covering}}}
& easy & - & 72.29 (14.0) & 300/300 & 300/300 \\
& medium & - & 64.48 (18.8) & 300/300 & 300/300 \\
& hard & 0.51 (1.3) & 27.02 (46.4) & 204/300 & 192/300 \\
\bottomrule
\bottomrule
\end{tabular}
\end{adjustbox}
\end{table}

\subsection{Performance of text-free separator configuration}\label{subsec:text_free}

Next, we evaluate the performance of our no-text pipeline (introduced in Section~\ref{sec:text_free_meth}) on a subset of MIPLIB 2017. By censoring descriptions and only providing a \texttt{.mps} file, we test whether LLM-based configuration can operate effectively without natural language input. For each MIPLIB instance, we generate up to five candidate problem descriptions based on the constraint types extracted from the \texttt{.mps} file. For each candidate description, we then produce 20 separator configurations using the pipeline from Section~\ref{subsec:generation}, yielding a configuration pool of up to 100 configurations for each instance. Because configurations are generated for a single instance, there is no natural validation dataset. As a result, we apply the ensembling strategy of LLM(0), which selects the medoid of the largest cluster, rather than more computationally intensive variants of our framework.

\begin{figure}[!tb]
    \centering
    \includegraphics[width=0.6\linewidth]{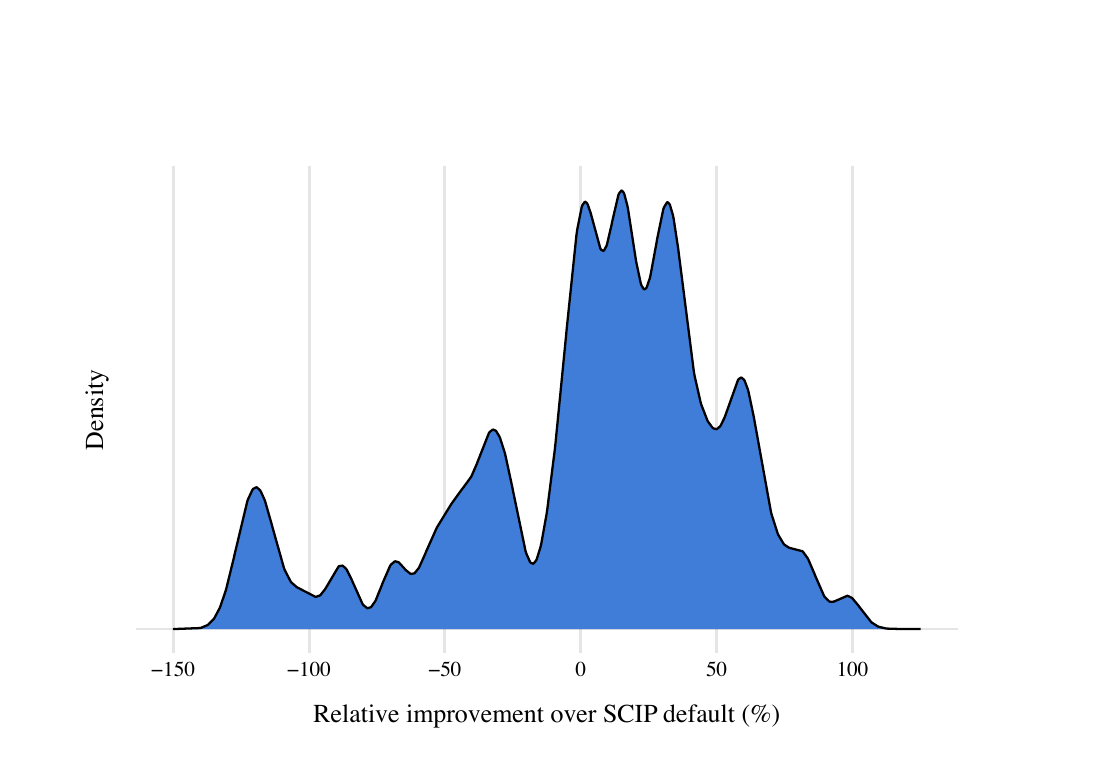}
    \caption{Distribution of relative improvements of LLM-no-text($0$) over SCIP defaults on filtered MIPLIB 2017 instances using the text-free framework.}
    \label{fig:text_free_scip}
\end{figure}

Figure~\ref{fig:text_free_scip} summarizes results by showing the distribution of relative improvements in solve time over SCIP. Our instance-specific configurations outperform the default separator settings, with a median improvement of 13.97\%. 
These results highlight that LLMs can help configure solvers directly from the \texttt{.mps} file, enabling zero-compute configuration at test time. However, consistent with observations by \citet{li2024learning}, the distribution of improvements exhibits a left skew: while many instances benefit substantially, instances where our LLM-based configuration fails tend to degrade performance by a larger margin. Future work could explore methods for identifying such instances in advance and reverting to solver defaults when appropriate.

\subsection{Ablations} \label{subsec: ablations}

\begin{table}
\centering
    \caption{Median (higher is better) and IQR of relative solve time improvements (\%) over default SCIP separator configuration for variants of our algorithm.}
    \label{tab:ablation}
    \begin{adjustbox}{max width=0.8\textwidth,center}
\begin{tabular}{p{5cm} cccc}
\toprule
 & \textbf{Ind. Set} & \textbf{Max. Cut} & \textbf{Bin. Pack.} & \textbf{MM} \\
\midrule
\textsc{LLM}(5) & 71.95 (5.88) & 71.01 (2.81) & 38.35 (7.99) & 53.3 (12.71) \\ \midrule \addlinespace[1mm]
Disable Cutting Planes & -14.96 (62.16) & 71.25 (11.8) & 30.43 (26.55) & -150 (86.12) \\  \addlinespace[1mm]
Different Problem Descr. & 23.4 (12.77) & 1.01 (8.49) & 16.07 (7.48) & -17.51 (13.14) \\ 
\midrule  \addlinespace[1mm]
\textbf{Prompt Components} & \\  \addlinespace[1mm]
No Sep. Text Descr. & 72.27 (5.82) & 71.49 (2.76) & 16.85 (9.52) & 9.29 (11.5) \\ \addlinespace[1mm]
No \LaTeX~Model & 72.45 (5.89) & 71.02 (2.87) & 14.55 (10.81) & 50.64 (11.38) \\ \addlinespace[1mm]
No Text Descr. & 41.37 (12.14) & 54.7 (4.52) & 18.15 (9.18) & -7.4 (17.8) \\ \addlinespace[1mm]
\midrule
\textbf{Ensemble Strategies} & \\  \addlinespace[1mm]
Average Configuration & 20.65 (13.43) & 71.24 (2.77) & 17.52 (9.5) & -11.08 (11.22) \\ \addlinespace[1mm]
Mode Configuration & 21.08 (13.43) & 71.44 (2.79) & 18.11 (9.31) & -12.63 (20.11) \\ \addlinespace[1mm]
Smallest Configuration & 20.83 (13.46) & 70.91 (2.86) & 17.42 (9.44) & -4.74 (17.75) \\ \midrule \addlinespace[1mm]
\textbf{$\mathcal{K}_\textup{eval}$ Size} & \\  \addlinespace[1mm]
$|\mathcal{K}_\textup{eval}|=1$ & 71.95 (5.88) & 71.01 (2.81) & 16.86 (7.39) & -23.27 (23.0) \\ \addlinespace[1mm]
$|\mathcal{K}_\textup{eval}|=5$ & 71.95 (5.88) & 71.01 (2.81) & 5.18 (5.06) & 9.5 (11.32) \\ \addlinespace[1mm]
$|\mathcal{K}_\textup{eval}|=20$ & 71.95 (5.88) & 71.01 (2.81) & 38.35 (7.99) & 53.3 (12.71) \\
\bottomrule
\end{tabular}
\end{adjustbox}
\end{table}

To gauge the importance of different elements of our approach on the configuration performance, we run a sequence of ablations on a representative sub-sample of our evaluation datasets, the results of which are included in Table \ref{tab:ablation}. 

First, we validate that our evaluation includes settings where cutting planes are important to the performance of the MILP solver. On two of the four datasets, disabling all cutting plane separators reduces performance compared to the SCIP default, while \textsc{LLM}(5) improves over the default. This observation confirms that \textsc{LLM}(5) can find non-trivial cutting plane configurations when cutting planes can improve solver performance. However, disabling all cutting planes has performance comparable to \textsc{LLM}(5) on the other two datasets. This observation highlights that in settings where cutting planes do not help performance, our approach finds lightweight cutting plane configurations that do not substantially slow down the solver.

We also validate that our LLM pipeline responds to the given problem description, rather than simply generating good generic configurations. To test this hypothesis, we generate configurations using a random problem description that differs from the description of the problem we evaluate on. Across all datasets, using a mismatched problem description leads to worse performance. 

Next, we evaluate the impact of different elements of the prompt used to generate the separator configurations, including (1) the text description of separators available, (2) the text description of the optimization model, and (3) the \LaTeX~description of the optimization model. Replacing the text descriptions with just the name of the cutting plane (e.g., \textit{Clique Cutting Planes}) leads to comparable performance on half of the datasets but a notable decrease in performance for both binary packing and middle-mile consolidation network design distributions. The latter are both settings where less well-known cutting plane families (e.g., Implied Bound and Flow Cover Cuts) are used in the final configuration, indicating that providing summarized information about the separators is important for  separators that are underrepresented in the LLM's training data. Both independent set and maximum cut problems employ well-studied separator families (e.g., clique cuts), which lends further credence to this claim. 

We also investigate the impact of different ensembling strategies over the same pool of candidate LLM configurations. Across all four datasets, ensembling the pool of configurations by averaging (i.e., rounding the average number of times each cutting plane is used over configuration pool), selecting the most common configuration, and taking the smallest configuration (i.e., the configuration that turns on the fewest cutting planes) yield similar performance to our cold-start approach that selects the medoid of the largest cluster. However, selecting the medoid with the best performance on the validation dataset $\mathcal{K}_\textup{val}$ leads to a significant improvement in performance over the other ensembling methods. 
Finally, we test the impact of the size of the validation dataset ${\cal K}_\textup{val}$ on the performance of the approach. As expected, adding more validation instances leads to better performance for half of the problem families, as small validation datasets provide a very noisy signal of the performance of a configuration on the evaluation dataset. However, both the independent set and maximum cut problems yield the same performance with only a single training instance.

\section{Conclusion}

In this paper, we introduced an LLM-based framework for cutting plane separator configuration in MILP solvers. To the best of our knowledge, this work is the first to explore the use of LLMs for algorithm configuration. Our framework leverages LLMs to generate individual separator configurations based on natural language descriptions of both the optimization problem and cutting plane separators. We then apply a $k$-medoids clustering algorithm to consolidate the generated configurations into a single robust final configuration. We also extend our approach to settings with no text description, and show that we can generate high-performing configurations based solely on the \texttt{.mps} file. Empirically, our framework demonstrates competitive performance improvements when benchmarked against state-of-the-art instance-agnostic methods for solver configuration. Importantly, it achieves these improvements while requiring significantly fewer computational resources. Additionally, our approach maintains strong performance on non-standard MILP formulations arising from real-world applications. We believe this is an intriguing demonstration of the potential for LLMs to augment traditional data used in existing ML methods with natural language data. This capability opens new avenues for integrating unstructured, descriptive information into algorithm configuration pipelines.

We consider this work an important first step towards LLM-based MILP solver configuration. Below, we outline several key challenges and potential directions for future exploration.

\paragraph{Instance-specific separator configuration:} While this work focuses on generating instance-agnostic separator configurations, tailoring configurations to specific instances could further improve performance. Combining LLM-generated configurations with existing portfolio-based algorithm configuration models like Hydra-MIP \citep{xu2011hydra} or L2Sep \citep{li2024learning} is an interesting direction for future research.


\paragraph{Smooth trade-offs between computation and performance:} Our framework provides significant computational savings compared to previous approaches. However, more work remains to refine the trade-off between computational cost and performance. For example, adaptive methods that dynamically adjust clustering parameters or the number of configurations generated could be developed to better balance resource usage and solver efficiency for specific use cases.

\paragraph{Fine-tuning with more domain knowledge:} Domain-specific knowledge could improve the quality of LLM-generated configurations. Generic LLMs, trained on broad internet data, are adept at addressing problems that are widely studied and discussed. However, their performance may falter in specialized domains due to a lack of fine-grained expertise. Fine-tuning LLMs using curated cutting plane research or other MILP-specific studies could help reduce noise in their outputs, align their recommendations more closely with domain best practices, and improve solver performance.



\bibliographystyle{style/informs2014} 
\bibliography{references} 

\section{Online Supplement}
In this online supplement, we outline additional details about the datasets used in our numerical experiments, and present additional results for configuring a leading commercial MILP solver, Gurobi \citep{achterberg2019gurobi}.

\subsection{Dataset details} \label{app:exp-details}

In this section we describe additional details about the datasets used in our numeric study. 

\paragraph{Distributional MIPLIB Instance Families.} Table \ref{table: data-description} provides an overview of the standard distributional MILP families used in our base experiments. The first seven families represent classic combinatorial optimization problems including maximum cut and set cover. \textit{Load Balancing}, featured in the ML4CO competition \citep{gasse2022machine} and categorized as \textit{hard} in the Distributional MIPLIB library, is a large-scale job-scheduling problem with side constraints that ensure robustness to machine failure. The \textit{Middle-Mile Consolidation Network Design} problem \citep{greening2023lead} determines a minimum-cost allocation of capacity on a transportation network involving vendors, fulfillment centers, and last-mile delivery centers. 

\paragraph{MIP-Evolve Dataset.}
MIP-evolve \citep{li2024foundation} is an LLM-powered framework for automatically generating diverse, complex MILP classes that fall out of distribution relative to commonly studied benchmarks. Unlike fixed, hand-curated datasets, MIP-evolve starts with an existing class of MILP and iteratively prompts the model to add, delete, and mutate constraints, and to cross over with other classes, while grounding each output class in plausible real-world contexts tailored to specific industry needs. A filtering stage then screens candidates for mathematical feasibility and practical usefulness.
Visual analyses of resulting problem classes in the space of t-Distributed Stochastic Neighbor Embedding (T-SNE) indicate that many generated classes occupy regions of feature space far from traditional MILP families, challenging methods to generalize beyond the usual problem types. 

\paragraph{MIPLIB 2017 Filtering.} To evaluate the performance of our text-free configuration pipeline, we benchmark our performance on a filtered set of instances from MIPLIB 2017 \citep{gleixner2021miplib}. We use the same filtering operation as \citet{li2024learning} and discard any instances that are infeasible, solved after presolving, or the primal-dual gaps are larger than 10\% after 300s of solve time. We also filter out instances that solve in under 1 second of work units on Gurobi.

\subsection{Additional results} \label{app:gurobi}
This section includes performance benchmarking against Gurobi default configurations for each of the experiments considered in the main text, as well as any omitted SCIP results.
\subsubsection{Standard distributional MILP benchmarks}
\begin{figure}[!t]
    \centering
        \begin{subfigure}
      \centering
      \includegraphics[width=0.45\textwidth]{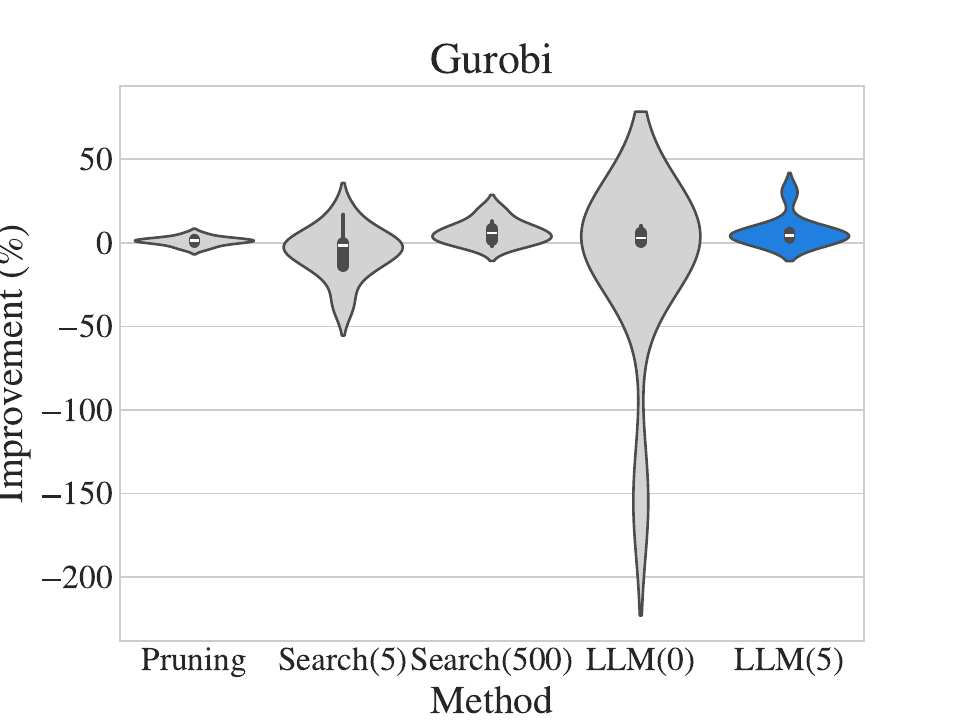}
      \includegraphics[width=0.45\textwidth]{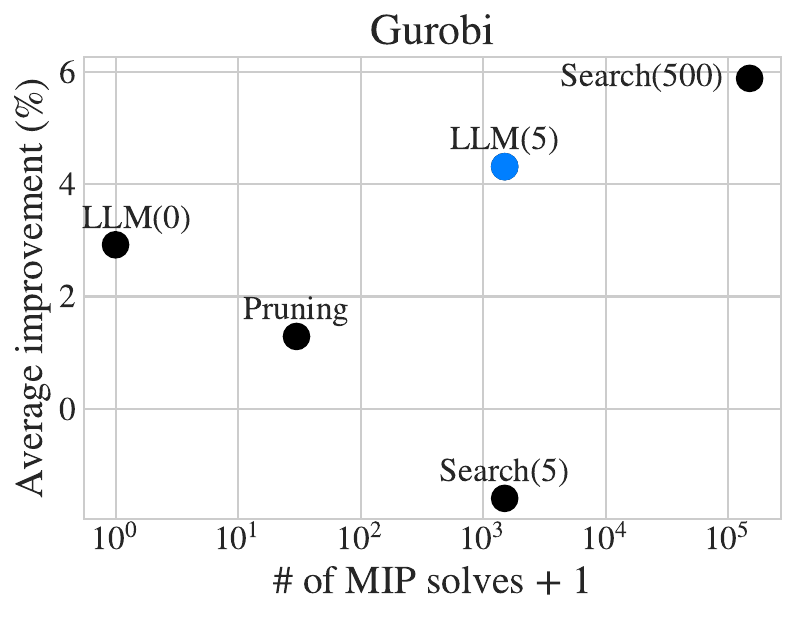}
    \end{subfigure}
    \caption{Distribution of median performance gain over evaluation datasets (higher is better) for all algorithms on Gurobi. Note that \textsc{LLM}(5) is able to achieve comparable performance to Search(500) at 1\% of the computational effort.}
    \label{fig: violin_plot_gurobi}
\end{figure}

Table \ref{table: gurobi-benchmark} reports the solve time improvements for separator configurations generated by our LLM-based algorithm and baselines compared to Gurobi defaults. Compared to SCIP, \textsc{LLM}(5) has more modest runtime improvements over Gurobi defaults but still outperforms the defaults between 0.6-30\%. Notably, \textsc{LLM}(0) still outperforms Gurobi defaults on all but two instance families. Figure \ref{fig: violin_plot_gurobi} summarizes the results of Table \ref{table: gurobi-benchmark}. It shows the distribution of improvements 
and highlights the trade-off between the number of MILPs solved and performance for each configuration method with Gurobi. Once again, our approach yields a new point on the Pareto frontier trading off computation time vs performance.

\begin{table}[!t] 
\centering
    \caption{Median (higher is better) and IQR of relative solve time improvements (\%) over default Gurobi separator configuration.}
    \label{table: gurobi-benchmark}
    \begin{adjustbox}{max width=\textwidth,center}
\begin{tabular}{
    @{} l
    @{\hspace*{12mm}}     c
    @{\hspace*{8mm}}     c
    @{\hspace*{8mm}}     c
    @{\hspace*{8mm}}     c
    @{\hspace*{8mm}}     c
                         c
    @{\hspace*{4mm}}     c
}
\toprule
 & \multicolumn{5}{c}{\textbf{Gurobi configuration}} \\
 \cmidrule(lr){2-6}
\textbf{Problem} & \textsc{Pruning} & \textsc{Search}(5) &  \textsc{Search}(500) & \textsc{LLM}(0) & \textsc{LLM}(5)\\
\midrule
\textbf{Bin. Pack.} & 0.53 (0.72) & 0.12 (2.34) & 12.94 (22.68) & 4.31 (5.56) & 4.31 (5.56) & \\\addlinespace[1mm]
\textbf{Cap. Fac.} & -1.82 (12.75) & -0.16 (15.12) & 8.37 (29.93) & 6.05 (45.41) & 6.05 (45.41) \\\addlinespace[1mm]
\textbf{Comb. Auc.} & 0.39 (7.44) & -1.59 (26.33) & 1.51 (25.45) & -1.05 (32.49) & 0.64 (26.39) \\\addlinespace[1mm]
\textbf{Ind. Set} & 1.83 (1.83) & -13.97 (11.66) & 7.87 (35.58) & 6.29 (39.72) & 6.33 (39.75) \\\addlinespace[1mm]
\textbf{Max. Cut} & 2.3 (0.93) & 16.84 (15.91) & 19.84 (14.92) & 10.17 (6.18) & 30.14 (15.0) \\\addlinespace[1mm]
\textbf{Pack.} & 5.16 (16.71) & -36.57 (27.27) & 5.88 (20.36) & 2.92 (17.37) & 2.92 (17.37) \\\addlinespace[1mm]
\textbf{Set Cov.}& 1.32 (1.64) & -2.33 (49.28) & 2.06 (40.74) & 1.77 (82.83) & 6.1 (43.31) \\\addlinespace[1mm]
\midrule
\textbf{Load Bal.} & 1.29 (2.09) & -0.8 (6.05) & 1.2 (15.22) & 0.41 (5.51) & 0.81 (5.6) \\\addlinespace[1mm]
\textbf{MM} & -3.65 (16.01) & -14.68 (20.12) & -2.06 (17.21) & -154.73 (4.78) & 4.3 (18.0) \\
\bottomrule
\end{tabular}
\end{adjustbox}
\end{table}

\subsubsection{Out-of-distribution instances}
Tables~\ref{tab:foundation_scip} and \ref{tab:foundational_gurobi} summarize the performance of our LLM-based approaches against random search baselines across problems from the MIP-evolve dataset \citep{li2024foundation}. Problem names reflect both the original base problem and any side constraints added during generation. For each problem type, we report the median solve time improvement (and IQR) over solver default configurations. As expected, the scale of improvements is more marginal for Gurobi than that of SCIP across many problem families we consider. Nevertheless, three observations stand out: first, \textsc{LLM}$(5)$ yields a modest improvement over Gurobi defaults for 24 of the 25 families. Second, the problem families with the worst relative performance are different when measuring with respect to SCIP vs Gurobi default configurations. This observation suggests that our model is not suffering from simple blind spots. Lastly, similar to our SCIP results, \textsc{LLM}$(5)$ matches or outperforms \textsc{Search}$(5)$ on 23 out of the 25 families with equal computational effort.  

For size generalization, table~\ref{table:gurobi-size} reports the performance of LLM$(5)$ relative to Gurobi defaults on MIPLIB problem families of varied size and difficulty. Performance is measured by relative improvement in solve time (\%, higher is better) when both LLM$(5)$ and the default configuration solve a problem to optimality, and by the additive difference in MIP gap (\%, higher is better) when both methods do not solve to optimality. While the improvements are more modest than those relative to SCIP defaults, the same general trends hold---performance varies by problem family and generally degrades with increasing problem difficulty.

\begin{table}[!tbh]
\centering
\caption{Performance on all easy MIP-evolve problems with a text description (SCIP).}
\label{table:milp-llm-performance}
\begin{adjustbox}{max width=\textwidth,center}
\begin{tabular}{
    @{} l
    @{\hspace*{12mm}} c
    @{\hspace*{8mm}} c
    @{\hspace*{8mm}} c
    @{\hspace*{8mm}} c
}
\toprule
 & \multicolumn{4}{c}{\textbf{SCIP configuration}} \\
\cmidrule(lr){2-5}
\textbf{Problem} & \textsc{Search}(5) & \textsc{Search}(500) & \textsc{LLM}(0) & \textsc{LLM}(5) \\
\midrule
Loyalty Rewards                 & 13.2\,(11.18) & 35.61\,(8.64)  & 9.75\,(12.09)   & 15.34\,(11.72) \\
Graph Coloring I               & 39.69\,(29.42) & 62.51\,(61.09) & 24.18\,(53.61)  & 44.29\,(26.81) \\
Graph Coloring II              & $-52.59$\,(47) & 42.14\,(25.47) & 12.37\,(62.92)  & 37.03\,(46.12) \\
Vessel Port Assignment         & $-150.35$\,(7.26) & 21.64\,(13.8) & 17.62\,(26.26)  & 45.93\,(16.27) \\
Complex Graph Coloring         & 19.94\,(45.25) & 77.63\,(27.5)  & 61.46\,(26.37)  & 75.27\,(28.13) \\
Job Shop Scheduling            & 14.27\,(69.97) & 88.06\,(14.27) & 87.93\,(17.4)   & 89.11\,(14.5) \\
Enhanced Multi-item Lotsizing  & $-108.2$\,(112.64) & $-20.66$\,(35.84) & $-115.37$\,(67.11) & $-73.45$\,(130.6) \\
Graph Coloring III             & $-4.65$\,(49.61) & 38.04\,(26.69) & 4.8\,(51.25)    & 32.25\,(39.94) \\
Job Scheduling                 & 16.97\,(37.76) & 21.78\,(54.48) & 0.86\,(84.27)   & 14.61\,(34.2) \\
Delivery Optimization          & $-39.9$\,(18.83) & 45.06\,(11.26) & 2.81\,(22.5)    & 40.87\,(11.75) \\
EV Placement                   & $-5.66$\,(37.45) & 61.49\,(9.29)  & 53.23\,(11.42)  & 61.65\,(11.8) \\
Job Scheduling II              & $-52.59$\,(47) & 35.99\,(41.05) & 33.4\,(44.17)   & 35.68\,(38.94) \\
Graph Coloring IV              & 36\,(59.93)    & 52.47\,(8.47)  & 12.71\,(21.77)  & 37.81\,(6.73) \\
Job Scheduling II              & $-58.02$\,(86.93) & 83.23\,(11.73) & 84.72\,(15.33)  & 84.72\,(15.33) \\
Vessel Port Assignment II      & $-1.53$\,(5.31) & 76.94\,(24.44) & 73.09\,(38.77)  & 73.09\,(38.77) \\
Graph Coloring V               & 13.21\,(40.06) & 58.82\,(18.75) & 11.19\,(45.55)  & 59.34\,(20.64) \\
Protein Folding                & 5.56\,(23.7)   & 36.39\,(16.38) & 23.66\,(41.13)  & 34.74\,(17.63) \\
Job Shop Scheduling II         & 98.81\,(1.61)  & 99\,(23.83)    & 76.71\,(26.01)  & 78.41\,(26.37) \\
Job Scheduling III             & 0.47\,(29.42)  & 82.56\,(19.81) & 81.23\,(22.14)  & 83.8\,(25.4) \\
Vehicle Routing                & $-70.88$\,(54.49) & 54.57\,(23.2) & $-23.82$\,(69.87) & $-23.82$\,(69.87) \\
Resource Allocation            & $-0.13$\,(31.16) & 50.82\,(19.55) & 42.57\,(22.38)  & 50.92\,(28.19) \\
Graph Coloring VI              & 1.69\,(14.87)  & 61.02\,(24.64) & 9.8\,(30.3)     & 26.31\,(25.6) \\
EV Placement II                & $-9.3$\,(46.8) & 63.89\,(13.97) & 54.78\,(21.98)  & 63.85\,(14.6) \\
Job Scheduling IV              & $-5.8$\,(86.93) & 83.46\,(20.23) & 86.26\,(14.43)  & 86.26\,(14.43) \\
Protein Folding II             & 28.62\,(18.96) & 92.39\,(1.65)  & 85.15\,(3.73)   & 85.15\,(3.73) \\
\bottomrule
\end{tabular}
\end{adjustbox}
\label{tab:foundation_scip}
\end{table}
\begin{table}[!t]
\centering
\caption{Performance on all easy MIP-evolve problems with a text description (Gurobi).}
\label{table:gurobi-milp-performance}
\begin{adjustbox}{max width=\textwidth,center}
\begin{tabular}{
    @{} l
    @{\hspace*{12mm}} c
    @{\hspace*{8mm}} c
    @{\hspace*{8mm}} c
    @{\hspace*{8mm}} c
}
\toprule
 & \multicolumn{4}{c}{\textbf{Gurobi configuration}} \\
\cmidrule(lr){2-5}
\textbf{Problem} & \textsc{Search}(5) & \textsc{Search}(500) & \textsc{LLM}(0) & \textsc{LLM}(5) \\
\midrule
Loyalty Rewards                 & $-30.57$\,(33.93) & 25.49\,(11.91) & 8.22\,(9.9)   & 8.35\,(11.15) \\
Graph Coloring I               & 0.86\,(7.32)   & 1.14\,(11.6)   & 0\,(18.13)    & 0\,(18.13) \\
Graph Coloring II              & 0.24\,(2.57)   & 2.12\,(18.91)  & 0.63\,(20.44) & 0.75\,(18.88) \\
Vessel Port Assignment         & 0\,(0.01)      & 4\,(2.85)      & 4.4\,(2.58)   & 4.41\,(2.67) \\
Complex Graph Coloring         & 0\,(0)         & 1.44\,(2.43)   & 0\,(3.23)     & 0\,(3.23) \\
Job Shop Scheduling            & 0\,(11.07)     & 23.86\,(21.06) & 12.98\,(24.94) & 12.98\,(24.94) \\
Enhanced Multi-item Lotsizing  & $-39.9$\,(18.83) & 26.62\,(48.78) & 0.71\,(38.1)  & 0.71\,(38.1) \\
Graph Coloring III             & $-5.66$\,(37.45) & 2.48\,(12.6)   & 1.68\,(18.66) & 1.68\,(18.66) \\
Job Scheduling                 & 0\,(7.63)      & 1.46\,(13.34)  & 0\,(8.7)      & 0\,(11.12) \\
Delivery Optimization          & 14.27\,(69.97) & 24.24\,(16.27) & $-9.96$\,(26.43) & 15.34\,(10.03) \\
EV Placement                   & $-0.01$\,(11.29) & 14.27\,(25.53) & 7.85\,(51.7)  & 7.85\,(51.7) \\
Job Scheduling II              & 1.0\,(6.89) &          3.09\, (6.61)   & 2.53\,(5.31)  & 3.19\,(6.58) \\
Graph Coloring IV              & $-12.39$\,(13.08) &  42.82\,(59.15)   & 23.5\,(53.33) & 23.5\,(53.33) \\
Job Scheduling II              & 11.78\,(31.97) & 26.74\,(28.21) & 20.8\,(26.36) & 21.17\,(26.29) \\
Vessel Port Assignment II      & $-3.3$\,(6.56) & 5.66\,(5.18)   & $-48.31$\,(90.59) & $-47.34$\,(92.44) \\
Graph Coloring V               & 0.38\,(18.65)  & 6.55\,(22.97)  & 9.95\,(24.19) & 10.32\,(23.87) \\
Protein Folding                & $-21.7$\,(62.33) & 44.46\,(28.56) & 27.69\,(24.9) & 27.69\,(24.9) \\
Job Shop Scheduling II         & $-4.45$\,(43.84) & 14.03\,(37.12) & 9.59\,(28.88) & 10.13\,(29.13) \\
Job Scheduling III             & $-12.94$\,(31.16) & 25.09\,(28.81) & 5.43\,(39.99) & 20.08\,(31.76) \\
Vehicle Routing                & $-30.3$\,(26.67) & 16.29\,(30.3)  & 12.72\,(30.78) & 24.4\,(26.01) \\
Resource Allocation            & $-10.47$\,(27.75) & 22.34\,(23.72) & $-3.41$\,(19.22) & 10.84\,(16.73) \\
Graph Coloring VI              & 0.05\,(30.22)  & 6.75\,(16.21)  & 7.6\,(23.05)  & 7.6\,(23.05) \\
EV Placement II                & $-3.73$\,(55.78) & 5.54\,(48.36)  & 7.7\,(53.72)  & 7.7\,(53.72) \\
Job Scheduling IV              & $-49.27$\,(67.84) & 27.4\,(28.75) & 21.32\,(28.44) & 21.32\,(28.44) \\
Protein Folding II             & 14.27\,(69.97) & 54.59\,(12.61) & $-3.69$\,(58.18) & 14.42\,(46.72) \\
\bottomrule
\end{tabular}
\end{adjustbox}
\label{tab:foundational_gurobi}
\end{table}

\begin{table}[!t]
\centering
\caption{Performance scaling across MIPLIB families of increasing size and difficulty (Gurobi).}
\label{table:gurobi-size}
\begin{adjustbox}{max width=\textwidth,center}
\begin{tabular}{llccccc}
\toprule
 & \multicolumn{5}{c}{\;\;\;\textbf{Performance metrics}} \\
\cmidrule(lr){3-6}
\textbf{Problem} & \textbf{Difficulty} &\;\; MIP gap difference (\%)$\uparrow$ \;\;& Improvement (\%)$\uparrow$ \;\; & LLM(5) \# solved  & Default \# solved \\
\midrule
\midrule
\multirow{3}{*}{\shortstack[l]{\textit{Combinatorial}\\\textit{auction}}} 

& very easy & - & 15.7 (31.7)   & 300/300 & 300/300 \\ 
& medium    & - & 6.0 (35.6)   & 300/300 & 300/300  \\
& very hard & 0.2 (0.5) & -        & 0/300   & 0/300   \\
\midrule
\multirow{2}{*}{\shortstack[l]{\textit{Capacitated facility}\\\textit{location}}} 
& easy & - & -2.8 (31.6)   & 300/300 & 300/300 \\
& medium & - & -0.1 (42.0)   & 300/300 & 300/300 \\
\midrule
\multirow{3}{*}{\shortstack[l]{\textit{Maximum}\\\textit{independent set}}}
& easy & - & 1.7 (2.9)   & 300/300 & 300/300 \\
& medium & 0.0 & 5.3 (19.6)   & 291/300 & 296/300 \\
& very hard & 4.0 (1.7) & -        & 0/300   & 0/300   \\
\midrule
\multirow{3}{*}{\shortstack[l]{\textit{Middle-mile}\\\textit{consolidation}\\\textit{network design}}}
& very easy  & -  & 6.6 (19.2)   & 300/300 & 300/300 \\
& medium  & - & -8.2 (45.1) & 300/300 & 300/300 \\
& hard & -0.8 (0.5) & -70.0 (152.1) & 52/300   & 96/300  \\
\midrule
\multirow{3}{*}{\shortstack[l]{\textit{Set covering}}}
& easy & - & 14.4 (18.7)   & 300/300 & 300/300 \\
& medium & -      & 13.7 (22.4)   & 300/300 & 300/300 \\
& hard & 0.0 (0.5) & 10.3 (25.4)   & 204/300 & 201/300 \\
\bottomrule
\bottomrule
\end{tabular}
\end{adjustbox}
\end{table}

\subsubsection{Text-free separator configuration}

\begin{figure}[!tb]
    \centering
    \includegraphics[width=0.8\linewidth]{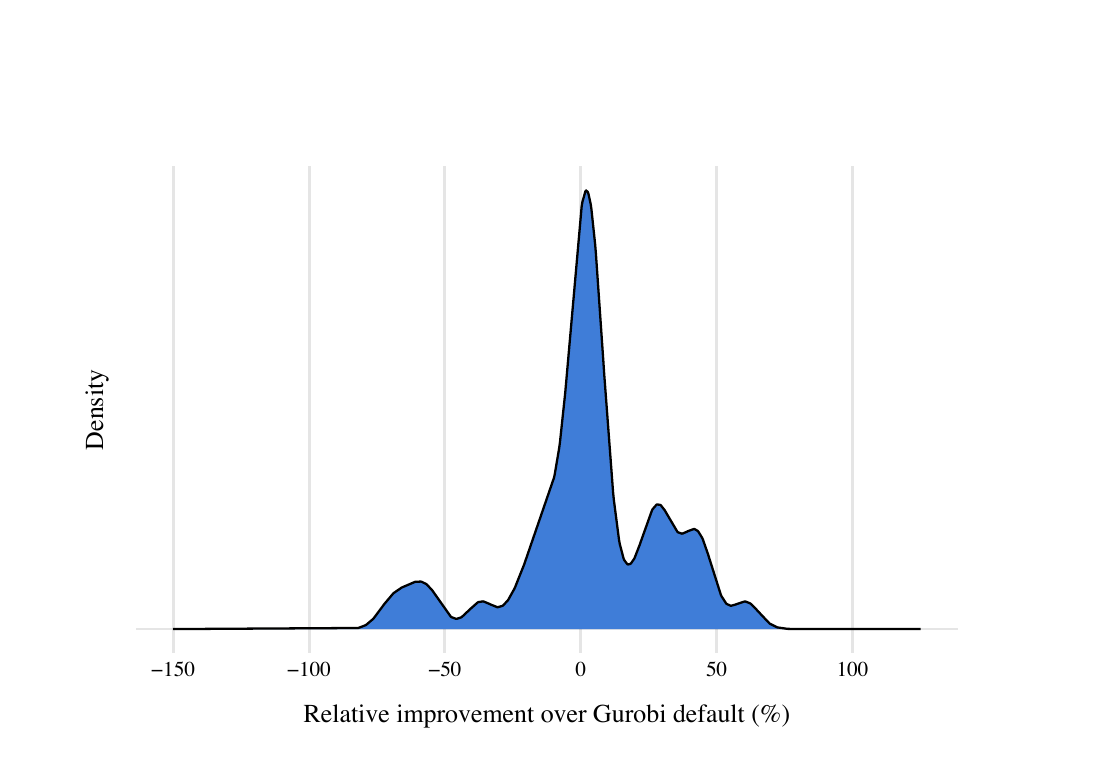}
    \caption{Relative improvement of LLM-no-text($0$) over Gurobi defaults on filtered MIPLIB 2017 instances using the text-free framework.}
    \label{fig:text_free_gurobi}
\end{figure}

Figure~\ref{fig:text_free_gurobi} summarizes results by showing the distribution of relative improvements in solve time over Gurobi. Similar to results on standard distributional benchmarks, our instance-specific configurations outperform the default separator settings by a more modest median improvement of 2.12\%.

\end{document}